\pdfoutput=1

\documentclass[11pt]{article}

\usepackage[final]{acl}

\usepackage{times}
\usepackage{latexsym}
\usepackage{tabularx}
\usepackage{color}
\usepackage{xcolor}
\usepackage{multirow}
\usepackage{colortbl}
\usepackage{amssymb}
\usepackage{pifont}
\usepackage{float}
\usepackage{enumitem}
\usepackage[T1]{fontenc}

\usepackage[utf8]{inputenc}

\usepackage{microtype}

\usepackage{inconsolata}

\usepackage{graphicx}

\definecolor{instructioncolor}{HTML}{fffea3}
\definecolor{exemplarcolor}{HTML}{8de5a1}
\newcommand{\cmark}{\ding{51}}%
\newcommand{\xmark}{\ding{55}}%

\title{LLMs Behind the Scenes: Enabling Narrative Scene Illustration}

\author{
\textbf{Melissa Roemmele$^1$\thanks{Correspondence: mroemmele@midjourney.com} \quad John Joon Young Chung$^1$ \quad Taewook Kim$^2$} \\\textbf{Yuqian Sun$^1$ \quad Alex Calderwood$^3$ \quad Max Kreminski$^1$} \\
$^1$Midjourney \quad $^2$Northwestern University \quad $^3$University of California, Santa Cruz\\
}

\begin{document}
\maketitle
\begin{abstract}
Generative AI has established the opportunity to readily transform content from one medium to another. This capability is especially powerful for storytelling, where visual illustrations can illuminate a story originally expressed in text. In this paper, we focus on the task of narrative scene illustration, which involves automatically generating an image depicting a scene in a story. Motivated by recent progress on text-to-image models, we consider a pipeline that uses LLMs as an interface for prompting text-to-image models to generate scene illustrations given raw story text. We apply variations of this pipeline to a prominent story corpus in order to synthesize illustrations for scenes in these stories. We conduct a human annotation task to obtain pairwise quality judgments for these illustrations. The outcome of this process is the \textsc{SceneIllustrations} dataset, which we release as a new resource for future work on cross-modal narrative transformation. Through our analysis of this dataset and experiments modeling illustration quality, we demonstrate that LLMs can effectively verbalize scene knowledge implicitly evoked by story text. Moreover, this capability is impactful for generating and evaluating illustrations.
\end{abstract}

\section{Introduction}\label{section:introduction}

Observing the transformation of a story from one modality to another (e.g. from text to visual form) can make the story more compelling to its audience. Recent advances in generative AI have enabled this kind of cross-modal transformation to be performed automatically. In particular, text-to-image models allow people to create visual material using natural language alone. Current interaction with these models typically involves users envisioning a particular visual target and then crafting language that realizes that target. Many stories that currently only exist in text form would be well-suited for transfer to an image modality, but the text itself of these stories may not be naturally optimal for directly applying text-to-image models. Given their demonstrated success at meta-prompting \citep[e.g.][]{zhou2023large}, large language models (LLMs) may be able to interface with story text to synthesize suitable prompts for text-to-image models towards this end. The cooperation between these AI models would make it possible to automatically generate illustrations for any given text-based story. 

\begin{figure}[h!]
\centering
\includegraphics[width=\linewidth]{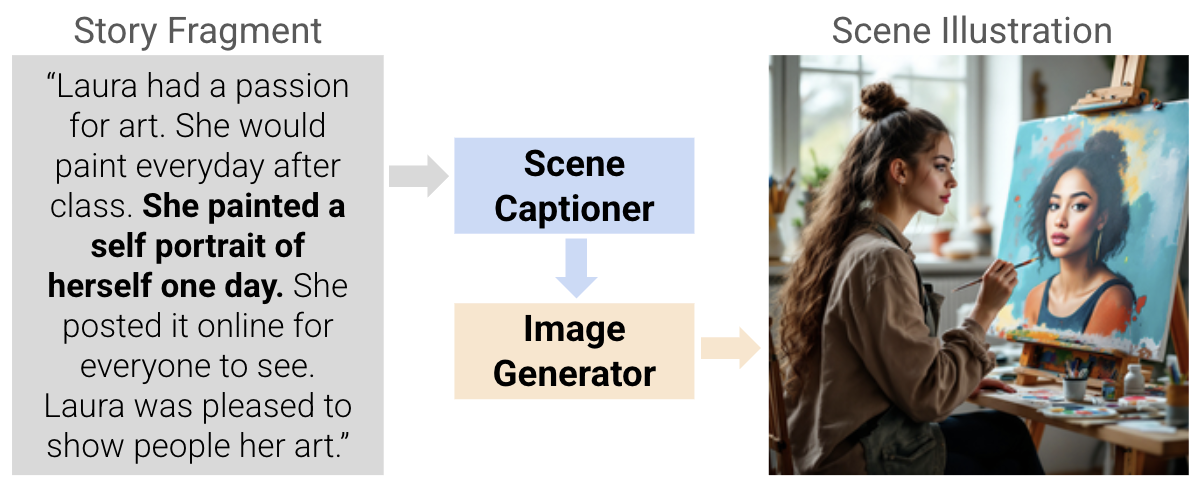}
\caption{Overview of scene illustration pipeline}
\label{figure:scene-illustration-pipeline}
\end{figure}

In this paper, we exemplify this approach to visual transfer of story text. Generating illustrations for stories, a task that has been termed \textit{story visualization}, encompasses a myriad of challenges. Some of these challenges pertain to modeling the relation between the story text and illustrations (text-image alignment), while others pertain to the relation between illustrations for different scenes in the story (image-image alignment). Existing story visualization research \cite[e.g.][]{li2019storygan} has largely focused on image-image alignment, in particular the problem of ensuring visual consistency between depictions of story elements like characters and settings. We aim to bring more research attention to issues of text-image alignment in this domain. Thus, our work is scoped to focus on individual \textit{scene illustrations}. In particular, we consider scene-level units of stories (\textit{fragments}). We present a pipeline (outlined in Figure \ref{figure:scene-illustration-pipeline}) that generates a scene illustration given a fragment in its story context. Through systematic variation and ablation of the components of this pipeline, we produce a novel set of scene illustrations for fragments in a notable dataset of stories, the ROCStories corpus \cite{mostafazadeh-etal-2016-corpus}. We then conduct a human annotation task to obtain relative quality judgments for pairs of illustrations. We refer to the resulting quality-annotated items as the \textsc{SceneIllustrations} dataset.

We leverage the \textsc{SceneIllustrations} dataset to demonstrate that LLMs can articulate visual knowledge of narrative scenes by inferring this knowledge directly from story text, without any visual input. We establish this capability through two findings. First, we show that LLMs are an effective interface for transforming story text into prompts that facilitate text-to-image models to produce illustrations. Second, we show that LLMs can verbalize scene characteristics in a way that is useful for evaluating the quality of illustrations. In particular, we demonstrate an approach to predicting human-favored illustrations among pairs in our dataset, through which we apply LLM-specified scene characteristics as evaluation criteria for scoring illustrations. The success of this approach relative to a criteria-ablated baseline further suggests the utility of LLMs for articulating scene knowledge that is implicitly conveyed by story text.

\paragraph{Contributions}
This paper makes the following contributions\footnote{Code and prompts for our experiments are available at: \href{https://github.com/roemmele/narrative-scene-illustration}{github.com/roemmele/narrative-scene-illustration}}:

\begin{itemize}[noitemsep,nolistsep,leftmargin=*]
\item We define and motivate the task of narrative scene illustration in relation to existing research on visually aligned storytelling.
\item We demonstrate a pipeline for producing scene illustrations for any given story text. The pipeline components are fully interchangeable and can be used with any LLM and text-to-image models.

\item We apply our pipeline to synthesize scene illustrations for existing stories and elicit human quality annotations for pairs of these illustrations, resulting in the newly created \textsc{SceneIllustrations} dataset\footnote{The \textsc{SceneIllustrations} dataset is available at:\\ \href{https://huggingface.co/datasets/roemmele/SceneIllustrations}{huggingface.co/datasets/roemmele/SceneIllustrations}}.

\item Through analysis of the quality annotations in \textsc{SceneIllustrations}, we show that LLMs are an effective interface between story text and text-to-image models in facilitating scene illustration.

\item We assess an approach to predicting these quality annotations that involves applying LLM verbalizations of scene characteristics as evaluation criteria. We discuss the evaluation results as additional evidence that LLMs can articulate visual scene knowledge inferred from story text. 
\end{itemize}

\section{Background and Related Work}\label{related-work}

\paragraph{Image-Aligned Story Data} Datasets that pair story text with images have emerged from research on visually grounded story generation, which involves writing a story given a sequence of images. Human authors have performed this task for existing media-sourced images \cite{halperin2023envisioning,huang-etal-2016-visual,hong-etal-2023-visual-writing}. For the reverse-direction task of story visualization, which involves generating a sequence of images given story text, some research has leveraged videos for data creation \cite{li2019storygan,tao2024coin}. Distinct frames of video are sampled as static images, while crowdsourced descriptions of frames are designated as the story text \cite{li2019storygan,maharana-bansal-2021-integrating,maharana2022storydalle}. A key design factor of all the above datasets is that the story text is authored specifically in response to the images, rather than originating in text form. We explore an alternative process for visually aligning narratives by synthesizing images for existing text-based stories.

\paragraph{Multimodal Storytelling Systems} In addition to datasets, there are increasing demonstrations of story visualization systems, as well as systems that generate story text and images in parallel, i.e. multimodal story generation \cite{an2024openleaf,koh2023generating,singh2023where,wan2024metamorpheus,yang2024seedstorymultimodallongstory}. While some models applied to these use cases have been trained end-to-end on the specialized datasets described above \cite{feng-etal-2023-improved,maharana-bansal-2021-integrating,tao2024coin}, researchers have also begun to leverage generically pretrained models to expand the scope of these systems to open-domain storytelling \cite{delima2024imaginingimagesaistorytelling,gong2023interactive,rakshit2024building}. We follow suit in leveraging a plug-and-play pipeline for scene illustration.

\paragraph{Meta-Prompting for Text-to-Image Models} One challenge with using generic models for story visualization is that the story text itself is not necessarily an optimal prompt for text-to-image models. In particular, this text tends to lack detailed visual descriptions (e.g. the physical appearance of story elements like entities and locations), which are considered essential when providing instructions to text-to-image models \cite{maharana2022storydalle}. Users of these models who have become skilled in writing prompts have done so largely through an iterative process of observing what prompt language yields desirable images \cite{don-yehiya-etal-2023-human}. Even with this skill, significant effort is required to manually compose a prompt that captures the intended visual features of the scene corresponding to a story fragment. Following the paradigm of meta-prompting \cite[e.g.][]{zhou2023large}, there is a variety of research on automated prompt optimization for text-to-image models \cite{brade2023promptify,feng2024promptmagician,hao2023optimizing,wang2024promptcharm}, some of which establishes the effectiveness of LLMs in facilitating this process \cite{lian2024llmgrounded}. Accordingly, recent story visualization work has used LLMs as an interface for deriving text-to-image prompts from story text. In particular, \citet{gong2023interactive} and \citet{he2024dreamstoryopendomainstoryvisualization} instructed GPT-4 to transform a story into a series of scene-level prompts intended as input to text-to-image models. It is presumed that these synthesized prompts are more visually descriptive than the story text and thus produce better images, but this has not been empirically validated. Thus, we address this opportunity in our work.

\paragraph{LLMs for Image Evaluation} Assessing the degree of semantic alignment between images and text is a prominent research endeavor, which has primarily involved measuring their similarity when projected into a shared embedding space \cite[e.g.][]{hessel-etal-2021-clipscore}. Because of their capacity for visually descriptive language, even unimodal (text-only) LLMs can contribute to this endeavor. For instance, several works have demonstrated the utility of unimodal LLMs for zero-shot visual recognition tasks \cite{li2023zeroshot,maniparambil2023enhancing,menon2023visual,pratt2023what}. This line of research has recently extended to eliciting visual knowledge from LLMs as a strategy for text-to-image evaluation \cite{lin2025evaluating,lu2023llmscore,hu2023tifa}. Encouraged by recent demonstrations of LLM-based evaluation in multimodal story generation \cite{an2024openleaf}, we pursue this method for evaluating scene illustrations.



\paragraph{Criteria-based Evaluation with LLMs} In NLP, criteria is a means of anchoring evaluation to certain objectives \cite{yuan-etal-2024-llmcrit}. With the rapidly expanding LLM-as-a-judge paradigm, this has evolved to the point where LLMs are not just applying human-authored criteria to assess text, but are also generating their own  criteria \cite{cook2024tickingboxesgeneratedchecklists}. We examine LLMs' capacity to generate evaluation criteria for the scene illustration task.

\section{Scene Illustration Pipeline}\label{section:scene-illustration-pipeline}

We first outline the high-level components\footnote{We ran all model components using APIs, which we specify here for each model. Unless otherwise indicated, we used the default inference parameters defined by the model's API.} of the illustration pipeline in this section, before describing their application in the next section.

\paragraph{Story Fragmentation} In our work, we consider a \textit{scene} to be an abstract unit of a story that can be distinctly illustrated by a single image. The story text that aligns to a scene is referred to as a \textit{fragment}. Thus, the first step of producing a scene illustration is to identify its source fragment. Recent work has validated the use of LLMs for the related task of segmenting events in narrative text \cite{michelmann2025large}. Accordingly, we utilize an LLM for this fragmentation task, by instructing it to explicitly annotate the boundaries of all fragments in a given story. Table A.\ref{table:fragmentation-prompt} shows the prompt we provide to the LLM to facilitate this, where the input contains the story text and the LLM is expected to generate the same text with brackets demarcating the left and right boundaries of each fragment, as demonstrated by the exemplars. We parse this output with a simple regular expression to gather the list of fragments.

\paragraph{Scene Descriptions} Once a fragment is identified, the fragment with its story context can then be mapped to a \textit{scene description}. A scene description is a verbalization of what should be illustrated in the image corresponding to the fragment. This text serves as the input to the text-to-image model used to produce the scene illustration. As described below in \S\ref{section:scene-illustration-dataset}, we consider different types of scene descriptions in order to evaluate the capability of LLMs to generate these descriptions. 

\paragraph{Image Generation} As mentioned, the scene descriptions are the inputs to a text-to-image model, referred to here as an \textit{image generator}. While we use the term `illustration' to describe the end-to-end process that yields an image depicting a scene, the output of this process (i.e. the image generator output) is also called an \textit{illustration}.

\section{\textsc{SceneIllustrations} Dataset}\label{section:scene-illustration-dataset}

Each item in the \textsc{SceneIllustrations} dataset consists of a fragment with its story context, along with two illustrations depicting the fragment. The illustrations vary based on their scene description and/or the image generator used to produce them. The dataset consists of 2990 items in total, which were created in two phases that we detail in this section: Phase 1 yielded 1777 items and Phase 2 yielded 1213 items. Table \ref{table:pair-statistics} presents a numerical summary of the items in the dataset, which will be explained by the following subsections. For both phases, we describe the pipeline for synthesizing and annotating the items, and then present analyses of the annotation results.

\subsection{Story Text}

Seeking out a story corpus suitable for the scene illustration task, we ultimately selected the ROCStories corpus \cite{mostafazadeh-etal-2016-corpus}, which has been widely used for storytelling-related research in NLP \cite[e.g.][]{brei-etal-2024-returning,kong-etal-2021-stylized,mu-li-2024-causal}. This choice of corpus was based on some key considerations. In particular, these English-language stories were authored to adhere to basic narrative structure in a tightly length-constrained format. In particular, each story consists of five sentences conveying ``a causally (logically) linked set of events involving some shared characters''. Thus, we can expect that stories are composed of distinct fragments that are each appropriately visualized as a scene illustration. Moreover, the stories are narrations of everyday experiences that can be interpreted according to commonsense knowledge. This knowledge is general enough it is likely to be familiar to the model components of our illustration pipeline.

\subsection{Phase 1 Pipeline Details}\label{section:phase-1-pipeline-details}

We applied the pipeline outlined in \S\ref{section:scene-illustration-pipeline} to produce an initial set of scene illustrations, which we refer to as \textit{Phase 1} data. As inputs to the pipeline, we used the first 50 stories in the ROCStories dev set.\footnote{The dev and test items in ROCStories are actually designated as the Story Cloze Test, where items have a specific format: each story consists of four sentences plus two alternative fifth sentences, where one is the `correct' story ending and the other is the 'incorrect' ending. For each item, we discarded the incorrect ending and appended the correct ending after the initial four sentences to form a single five-sentence story.}

\paragraph{Fragmentation} We divided these stories into fragments as described in \S\ref{section:scene-illustration-pipeline}, using \textsc{claude-3.5}\footnote{claude-3-5-sonnet-20240620, ran via the \href{https://www.anthropic.com/api}{Anthropic API}} as the LLM. We selected \textsc{claude-3.5} because it topped the LLM Creative Story Writing Benchmark\footnote{\href{https://github.com/lechmazur/writing}{github.com/lechmazur/writing}}, which assesses story generation capabilities. As shown in Table A.\ref{table:fragmentation-statistics}, this resulted in 206 total fragments across all 50 stories, an average of 4.12 per story. \S \ref{phase-1-fragmentation-analysis} presents some additional analysis.

\paragraph{Scene Descriptions} We applied an LLM to transform a fragment alongside its story context into a scene description, using the prompt in Table A.\ref{table:scene-captioning-prompt} with \textsc{claude-3.5} as the LLM. We employ the term \textit{scene captioner} to refer to an LLM's role when running this prompt, and we refer to the outputs as \textsc{Caption} scene descriptions. As outlined in Table \ref{table:scene-description-types}, \textsc{Caption} is one of four scene description types we consider for Phase 1. We compare \textsc{Caption} with baseline scene descriptions composed of the raw story text. In the first baseline case, \textsc{NC-Fragment} (i.e. \underline{n}o \underline{c}ontext fragment), we use the original fragment isolated from its story context as a scene description. The obvious limitation of \textsc{NC-Fragment} is that the ablated context may be necessary for understanding certain information in the fragment (for example, a fragment might use a pronoun whose referent is only specified in the context). Thus, we considered two additional baseline scene descriptions that account for the story context, referred to as \textsc{VC-Fragment} and \textsc{SC-Fragment}. As Table \ref{table:scene-description-types} shows, \textsc{VC-Fragment} (i.e. \underline{v}erbose \underline{c}ontext) inserts the full story text into the scene description, which is formatted as an instruction to consider this context when illustrating the fragment. Alternatively, \textsc{SC-Fragment} (i.e. \underline{s}uccinct \underline{c}ontext) is a rewritten version of the fragment where references to information in the story context are made explicit, enabling the fragment to be understood independently of the context. We prompt an LLM (also \textsc{claude-3.5}) to do this rewriting task, using the prompt in Table A.\ref{table:fragment-rewriting-prompt}. Table \ref{table:scene-description-types} gives examples of these different scene descriptions, with additional examples in Table A.\ref{table:illustration-examples-phase-1}.

\begin{table*}
\small
\centering
\rowcolors{1}{white}{gray!20}
\begin{tabularx}{\textwidth}{p{0.135\textwidth} p{0.17\textwidth} X}
\textbf{Type} & \textbf{Format} & \textbf{Example}\\
\hline
\textsc{NC-Fragment} & \texttt{\{\{fragment\}\}} & ``Alice called her mother and apologized profusely.''\\
\textsc{VC-Fragment} & ``Consider this story: [\texttt{\{\{story\}\}}] Based on this context, illustrate this fragment of the story: [\texttt{\{\{fragment\}\}}]'' & ``Consider this story: [Alice was getting married in a few weeks. One night, her mother called and she forgot to call her back. Her mother left an angry message on her phone. She threatened not to come to the wedding. Alice called her mother and apologized profusely.] Based on this context, illustrate this fragment of the story: [Alice called her mother and apologized profusely.]''\\
\textsc{SC-Fragment} & LLM output of fragment rewriting prompt (Table A.\ref{table:fragment-rewriting-prompt}) & ``The bride-to-be called her mother and apologized profusely for forgetting to return her call and for the resulting angry message threatening not to attend the wedding.''\\
\textsc{Caption} & LLM output of scene captioning prompt (Table A.\ref{table:scene-captioning-prompt}) & ``A young woman with a worried expression sits on a couch, holding a phone to her ear. She's gesticulating with her free hand, appearing to speak emphatically. In the background, a wedding dress can be seen hanging on a closet door. The room is dimly lit, suggesting it's evening, and there's a notepad with wedding plans visible on a nearby coffee table.''\\
\hline
\end{tabularx}
\caption{Types of scene descriptions for Phase 1}
\label{table:scene-description-types}
\end{table*}

\paragraph{Image Generation} We then applied two image generators\footnote{With exception to Midjourney, we ran all image generation models described in this paper via the \href{https://replicate.com/}{Replicate API}.} to generate images using the scene descriptions as prompts. In particular, we used Midjourney v6.1, denoted here as \textsc{mj-6.1} \cite{midjourney-6.1}, and FLUX-1[pro], denoted here as \textsc{flux-1-pro} \cite{blackforestlabs-flux-1-pro}. We selected these image generators because they topped the Artificial Analysis Image Arena Leaderboard\footnote{\href{https://artificialanalysis.ai/text-to-image/arena/leaderboard-text}{artificialanalysis.ai/text-to-image/arena/leaderboard-text}} at the time of Phase 1 in August 2024. This leaderboard captures the relative ELO score \cite{boubdir-etal-2023-elo} of text-to-image models based on pairwise human judgments regarding how well images from different models reflect the input prompt. Table A.\ref{table:illustration-examples-phase-1} includes examples of generated illustrations.

\subsection{Phase 1 Annotation Task}\label{section:phase-1-annotation-task}

\paragraph{Illustration Pairs} Our primary objective for Phase 1 was to assess the effectiveness of the LLM-based scene captioner in generating illustrations relative to generating them directly from the raw story text. To address this, we randomly sampled pairs of illustrations each belonging to the same fragment (across 206 possible fragments), where one illustration used a \textsc{Caption} as the scene description, while the other used one of the baseline scene descriptions: \textsc{NC-Fragment}, \textsc{VC-Fragment}, or \textsc{SC-Fragment}. This sampling resulted in some pairs where the illustrations used the same image generator and others that used different image generators. Ultimately there were 1777 illustration pairs. Table \ref{table:pair-statistics} specifies their exact distribution. 

\begin{table}[h]
\centering
\small
\begin{tabular}{l c c}
\textbf{Illustration Pair Type} & \textbf{\# Pairs} & \textbf{$\kappa_u$}\\
\hline
\multicolumn{3}{c}{\underline{\textit{Phase 1}}}\\
All & 1777 & 0.436\\
\rowcolor{gray!20} Different Scene Descriptions & 1457 & 0.483\\
\quad\textsc{NC-Fragment} vs. \textsc{Caption} & 680 & 0.520\\
\quad\textsc{VC-Fragment} vs. \textsc{Caption} & 384 & 0.504\\
\quad\textsc{SC-Fragment} vs. \textsc{Caption} & 393 & 0.398\\
\rowcolor{gray!20} \multicolumn{3}{l}{Different Image Generators} \\
\quad\textsc{flux-1-pro} vs. \textsc{mj-6.1} & 661 & 0.364\\
\hline
\multicolumn{3}{c}{\underline{\textit{Phase 2}}}\\
All & 1213 & 0.231\\
\rowcolor{gray!20} Different Scene Captioners & 807 & 0.239\\
\quad\textsc{claude-3.5} vs. \textsc{gpt-4o} & 265 & 0.200\\
\quad\textsc{claude-3.5} vs. \textsc{llama-3.1} & 267 & 0.239\\
\quad\textsc{gpt-4o} vs. \textsc{llama-3.1} & 275 & 0.274\\
\rowcolor{gray!20} Different Image Generators & 809 & 0.231\\
\quad\textsc{flux-1.1-pro} vs. \textsc{ideogram-2.0} & 72 & 0.079\\
\quad\textsc{flux-1.1-pro} vs. \textsc{mj-6.1} & 74 & 0.183\\
\quad\textsc{flux-1.1-pro} vs. \textsc{recraft-v3} & 75 & 0.089\\
\quad\textsc{flux-1.1-pro} vs. \textsc{sd-3.5-large} & 70 & 0.183\\
\quad\textsc{ideogram-2.0} vs. \textsc{mj-6.1} & 98 & 0.314\\
\quad\textsc{ideogram-2.0} vs. \textsc{recraft-v3} & 81 & 0.159\\
\quad\textsc{ideogram-2.0} vs. \textsc{sd-3.5-large} & 94 & 0.339\\
\quad\textsc{mj-6.1} vs. \textsc{recraft-v3} & 72 & 0.426\\
\quad\textsc{mj-6.1} vs. \textsc{sd-3.5-large} & 87 & 0.189\\
\quad\textsc{recraft-v3} vs. \textsc{sd-3.5-large} & 86 & 0.271\\
\hline
\hline
\multicolumn{3}{c}{\underline{\textit{Phase 1 \& 2}}}\\
All & 2990 & 0.352\\
\end{tabular}
\caption{Illustration pair statistics for the \textsc{SceneIllustrations} dataset, divided into Phase 1 and Phase 2, and including inter-annotator agreement ({$\kappa_u$}) for different pair types. For the total number of unique illustrations associated with each scene description type and image generator, see Table A.\ref{table:unique-illustration-statistics}.}
\label{table:pair-statistics}
\end{table}

\paragraph{Task Design} We designed an annotation task to assess the relative quality of the two illustrations in each pair. In judging a pair, human annotators were shown the full story with the target fragment for that scene underlined, along with the two alternative images. Note that scene descriptions were not shown to annotators, since their judgment of illustration quality should be conditioned on the original text. As shown in Figure A.\ref{figure:annotation-interface}, annotators were instructed to select the image that was ``the better visualization of the underlined fragment''. Annotators could express uncertainty by selecting ``I can't decide which is better''. We implemented the UI for this task using \textsc{Potato} \cite{pei-etal-2022-potato}.

\paragraph{Procedure} We deployed the task on \href{https://www.prolific.co/}{Prolific} to obtain annotators. English proficiency was the only requirement for participation. We sought 2 annotators to judge each illustration pair. Each participant judged between 33 and 74 pairs (median=47), plus 3 ``attention check'' items where one illustration in the pair was replaced with one for a different story, making it trivially easy which image to select. Participants were paid \$6 for an expected completion time of 30 minutes. We filtered out participants who did not pass all of the attention check items. Ultimately, 75 (out of 80) participants passed the attention checks. This resulted in a total of 3554 responses for the 1777 pairs, where each item received a response from 2 annotators.

\subsection{Phase 1 Annotation Results}\label{phase-1-annotation-results}

\paragraph{Inter-annotator Agreement} Given the annotated pairs resulting from \S\ref{section:phase-1-annotation-task}, we computed the inter-annotator agreement of which illustration was selected as the better one in each pair. We did this using an \textit{uncertainty-weighted} variation of Cohen's Kappa score \cite{cohen1960coefficient}, which we abbreviate here as  $\kappa_u$. This variation considers that response disagreements arising from one annotator expressing uncertainty (i.e. selecting ``I can’t decide'') should be weighted half as much as disagreements where the two annotators each select a different illustration as better. As shown in Table \ref{table:pair-statistics}, the overall $\kappa_u$ for all 1777 items was 0.436, which can be classified as moderate agreement \cite{landis1977measurement}. Annotators agreed in their responses for 62.3\% of items. Table \ref{table:pair-statistics} also shows that agreement was higher in Phase 1 among the 1457 pairs where the illustrations used different scene descriptions ($\kappa_u$=0.483), while agreement was lower among the 661 pairs where the illustrations used different image generators ($\kappa_u$=0.364). This indicates that the scene description was particularly influential to annotators' judgments of relative illustration quality. 

\paragraph{Win Rates for Scene Description Types} To determine whether using an LLM as a scene captioner helps illustration quality, we counted how often the favored illustration was associated with each scene description type, i.e. each type's \textit{win rate}. Table \ref{table:annotation-task-1-scene-description-win-rates} shows the win rate for \textsc{Caption} illustrations when alternatively paired with \textsc{NC-Fragment}, \textsc{VC-Fragment}, and \textsc{SC-Fragment} illustrations. This win rate is represented as the percentage of responses in which annotators selected the \textsc{Caption} illustration as better among all responses for each respective set of pairs. In all three cases, the \textsc{Caption} is significantly\footnote{Statistical significance was computed using a one-sample binomal test at $\alpha=0.05$ to determine if the win rate was higher than that expected by chance, where chance is defined as $(1 - \# ties / \# responses) / 2$} better: it has an overall win rate of $\approx$78\% against \textsc{NC-Fragment}, $\approx$75\% against \textsc{VC-Fragment}, and $\approx$73\% against \textsc{SC-Fragment}. Table A.\ref{table:annotation-task-1-scene-description-win-rates-extended} further examines the win rates for pairs that used the same image generator, verifying that \textsc{Caption} is equally favorable regardless of which image generator is used. Considered along with the inter-annotator agreement results highlighted above, which showed higher agreement among pairs where the illustrations used different scene descriptions, we can specifically conclude that ablating the scene captioner (i.e. using the baseline \textsc{NC-Fragment}, \textsc{VC-Fragment}, or \textsc{SC-Fragment} scene descriptions) yielded illustrations that annotators consistently judged as lower quality relative to those that used the scene captioner. This validates the importance of using an LLM for scene captioning in the pipeline: the resulting verbalization enables the image generator to better depict how a story fragment should be visually illustrated as a scene.

\begin{table}[h!]
\small
\centering
\rowcolors{1}{white}{gray!20}
\begin{tabular}{l c c }
\textbf{Scene Description Pair} & \textbf{\textsc{Caption} Win \%}\\
\hline
\textsc{Caption} vs. \textsc{NC-Fragment} & 78.1\\
 \textsc{Caption} vs. \textsc{VC-Fragment} & 74.7\\
\textsc{Caption} vs. \textsc{SC-Fragment} & 72.5\\
\hline
\end{tabular}
\caption{Win rates of \textsc{Caption} over the baseline scene descriptions in Phase 1}
\label{table:annotation-task-1-scene-description-win-rates}
\end{table}


\subsection{Phase 2 Motivation and Design}\label{phase-2-motivation-and-design}

After observing that the \textsc{Caption} scene descriptions significantly contribute to illustration quality, we wanted to compare the impact of different LLMs as scene captioners. Phase 1 only considered \textsc{claude-3.5}. In \textit{Phase 2}, we included other LLMs that obtained noteworthy performance on the LLM Creative Story Writing Benchmark: \textsc{gpt-4o}\footnote{gpt-4o-2024-05-13, ran via the  \href{https://platform.openai.com/}{OpenAI API}} \cite{openai2024gpt4ocard} and \textsc{llama-3.1}\footnote{llama-3.1-405b-instruct, ran via the Replicate API} \cite{grattafiori2024llama3herdmodels}. We used the same captioning prompt from \S\ref{section:scene-illustration-dataset} (Table A.\ref{table:scene-captioning-prompt}). 

We expanded the Phase 2 data to include a larger set of fragments compared with those of Phase 1. We randomly sampled 1000 stories from the ROCStories dev set, split them into fragments using the same method from Phase 1 (\textsc{claude-3.5} with the Table A.\ref{table:fragmentation-prompt} prompt), then randomly selected one fragment per story for inclusion in the dataset. 

We also considered a larger set of image generators in Phase 2. Based on the state of the Artificial Analysis Leaderboard in November 2024, we selected five image generators. This included \textsc{mj-6.1} from Phase 1, as well as FLUX1.1[pro] (referred to here as \textsc{flux-1.1-pro}) \cite{blackforestlabs-flux-1.1-pro}, Ideogram 2.0 (\textsc{ideogram-2.0}) \cite{ideogram-2.0}, Recraft V3 (\textsc{recraft-v3}) \cite{recraft-v3}, and Stable Diffusion 3.5 Large (\textsc{sd-3.5-large}) \cite{stable-diffusion-3.5-large}.

We applied the scene illustration pipeline to produce illustrations for all 1000 fragments, varying runs of the pipeline between the 3 scene captioners and 5 image generators. We sampled a roughly equal ratio of pairs where the illustrations varied by scene captioner, image generator, or both scene captioner and image generator. The exact distribution is specified in Table \ref{table:pair-statistics}. We repeated the same procedure detailed in \S\ref{section:phase-1-annotation-task} to obtain selections from two annotators for the better illustration in each pair. There were 48 (out of 49 total) annotators on Prolific who passed the attention checks, each annotating between 46 and 109 pairs (median=50), resulting in a total of 2426 responses for 1213 pairs.

\subsection{Phase 2 Annotation Results}\label{phase-2-annotation-results}

\paragraph{Inter-annotator Agreement} As shown in Table \ref{table:pair-statistics}, the overall $\kappa_u$ for all 1213 items in Phase 2 was 0.231, and annotators agreed in their responses for 52.6\% of these items. This is lower than the overall agreement observed for Phase 1. 
Table \ref{table:pair-statistics} also shows that the agreement level was similar between the 807 pairs where illustrations involved different scene captioners ($\kappa_u$=0.239) and the 809 pairs that involved different image generators ($\kappa_u$=0.231). Agreement varied especially widely based on which particular image generators were paired together (ranging from 0.079 for \textsc{flux-1.1-pro} vs. \textsc{ideogram-2.0}, up to 0.426 for \textsc{mj-6.1} vs. \textsc{recraft-v3}). This indicates that in contrast to Phase 1 where there was a significant variable (the presence/absence of the scene captioner) that made the relative quality of illustrations more consistently distinguishable to annotators, the Phase 2 pairs were less reliably distinct.

\paragraph{Win Rates for Scene Captioners} Table \ref{table:annotation-task-2-scene-captioner-win-stats} shows the win rates for each LLM scene captioner against each of the others. In particular, each value is the percentage of responses where the illustration associated with the scene captioner in the row label was selected as better than the illustration associated with the scene captioner in the column label. Statistically significant win rates are denoted with an asterisk. Recall that a response of ``I can't decide'' indicates a tie, which is why win rates of less than 50\% may be statistically significant. These results show that \textsc{claude-3.5} yields the highest win rates, followed by \textsc{gpt-4o}, with  \textsc{llama-3.1} having lowest rates. The win rate for \textsc{claude-3.5} against \textsc{llama-3.1} is statistically significant, suggesting that the former generates more descriptive captions compared with the latter.

\begin{table}[h]
\small
\centering
\rowcolors{1}{white}{gray!20}
\begin{tabular}{l | c | c | c}
& \textsc{claude-3.5} & \textsc{gpt-4o} & \textsc{llama-3.1} \\
\hline
\textsc{claude-3.5} & - & 46.2 & 49.6*\\
\textsc{gpt-4o} & 41.1 & - & 48.0\phantom{*}\\ 
\textsc{llama-3.1} & 39.7 & 42.9 & - \\
\hline
\end{tabular}
\caption{Win rates (\%) by scene captioner for Phase 2}
\label{table:annotation-task-2-scene-captioner-win-stats}
\end{table}

\paragraph{Win Rates for Image Generators} While not the focus of our analysis, we observed some significant differences in the win rates of different image generators. These results appear in \S\ref{win-rates-image-generators}.

\begin{table*}[h!]
\small
\rowcolors{1}{white}{gray!20}
\begin{tabularx}{\textwidth}{m{0.7\textwidth} m{0.11\textwidth} m{0.11\textwidth}}
\hiderowcolors
\textbf{Fragment} (within story) & \textbf{Illustration 1} & \textbf{Illustration 2}\\
\hline
\rowcolor{white}Sophie's nana was terminally ill. Sophie visited her in the hospital to say goodbye. \textbf{Her nana gave Sophie her prized gold locket. She told Sophie to keep it to remember her by.} Sophie cried. & \includegraphics[width=0.11\textwidth]{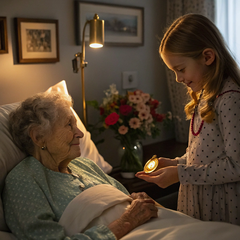} & \includegraphics[width=0.11\textwidth]{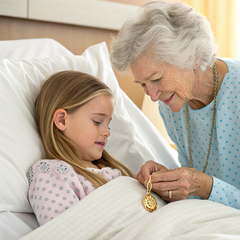}\\
\end{tabularx}
\begin{tabular}{p{0.7\textwidth} >{\centering}m{0.11\textwidth} >{\centering\arraybackslash}m{0.11\textwidth}}
\hline
\textbf{Criteria} & \textbf{Response} & \textbf{Response}\\
\hline
\end{tabular}
\small
\begin{tabular}{p{0.7\textwidth} >{\centering}m{0.11\textwidth} >{\centering\arraybackslash}m{0.11\textwidth}}
\showrowcolors
1. The image shows two people: an elderly woman (nana) and a younger woman (Sophie) & \cmark & \xmark\\
2. The setting appears to be a hospital room or medical facility & \cmark & \cmark\\
3. The elderly woman is in a hospital bed or medical chair & \cmark & \xmark\\
4. The image shows a gold locket & \cmark & \cmark\\
5. The locket is clearly visible and recognizable as a piece of jewelry & \cmark & \cmark\\
6. The elderly woman is holding or presenting the locket to the younger woman & \xmark & \cmark\\
7. The younger woman's hand is positioned to receive or touch the locket & \cmark & \xmark\\
8. The facial expressions of both women convey emotional significance & \cmark & \cmark\\
9. The elderly woman's expression shows love, tenderness, or sadness & \cmark & \cmark\\
10. The younger woman's expression shows a mix of emotions (sadness, gratitude, love) & \cmark & \xmark\\
11. The body language of both women suggests intimacy and connection & \cmark & \cmark\\
12. The composition focuses on the moment of giving/receiving the locket & \cmark & \cmark\\
13. The lighting adequately illuminates the locket and the faces of both women & \cmark & \cmark\\
14. The locket appears to be in good condition, suggesting its value as a keepsake & \cmark & \cmark\\
15. The elderly woman's appearance suggests illness or frailty & \cmark & \xmark\\
16. The younger woman's appearance and demeanor suggest she is visiting & \cmark & \xmark\\
17. The overall atmosphere of the image conveys a solemn and meaningful moment & \cmark & \cmark\\
18. The spatial relationship between the two women suggests closeness and care & \cmark & \cmark\\
19. Any medical equipment or hospital elements are present but not dominating the scene & \cmark & \cmark\\
20. The perspective allows viewers to see both the locket and the emotional exchange between the women & \cmark & \cmark\\
\hline
\end{tabular}
\small
\begin{tabular}{p{0.7\textwidth} >{\centering}m{0.11\textwidth} >{\centering\arraybackslash}m{0.11\textwidth}}
\rowcolor{gray!40} & Score=19.0 & Score=14.0\\
\end{tabular}
\caption{Demonstration of criterial rating approach applied to both illustrations in a given pair.  In this particular example, the criteria writer is \textsc{claude-3.5}, and the rater providing each response is \textsc{gpt-4o}.}
\label{table:criteria-scoring-example}
\end{table*}

\section{Predicting Illustration Quality}

\subsection{\textit{Perfect-Agreement} Data Subset}

The \textsc{SceneIllustrations} dataset provides an opportunity to understand what characterizes a successful transformation of a narrative scene from text to image form. To initiate this line of work, we explored a particular approach to modeling annotators' judgments of relative illustration quality. For this experiment, we combined the items from Phase 1 and Phase 2, and disregarded items involving annotator disagreement. The resulting \textit{Perfect-Agreement} subset consists of 1745 items ($\approx$58\% of the full dataset) where both annotators agreed in their selection of the better illustration in the pair.

\subsection{Criteria Generation}\label{section:criteria-generation-approach}
Our approach leverages the finding from \S\ref{section:scene-illustration-dataset} that LLMs can effectively verbalize visual descriptions of scenes based on the story text. We consider whether these descriptions can be used as \textit{criteria} for predicting illustration quality. For each fragment, we ran the prompt in Table A.\ref{table:criteria-generation-prompt} to produce criteria articulating the expected visual characteristics of the scene illustration. We use the term \textit{criteria writer} to refer to an LLM's role when running this prompt, and we refer to its output as a \textit{criteria set}. An example of a criteria set is included in Table \ref{table:criteria-scoring-example}. Note that a criteria writer model does not require vision capabilities, since it observes only the story text as input. 

Two design considerations for the criteria generation prompt were \textit{flexibility} and \textit{atomicity}. Flexibility emphasizes that a scene characteristic referenced by a criterion may be depicted with multiple alternative visual details that all align acceptably well with the story text. For example, if a criterion conveys that the scene should take place at a particular location, it should be flexible about how the location is portrayed. Regarding atomicity, we aimed for each criterion to be as atomic as possible, meaning that it should refer to only a single characteristic of the scene. This promotes concise and easy-to-parse responses when judging whether the criterion is satisfied by an image, as opposed to a criterion that conflates multiple characteristics, some of which are satisfied and others that are not. The prompt did not specify a particular number of criteria to return, but it indicated that the criteria set should comprehensively refer to as many scene characteristics as possible without redundancy.

\paragraph{Criteria Writer Details} We examined three criteria writers, the same LLMs that operated as scene captioners in \S\ref{phase-2-motivation-and-design}: \textsc{claude-3.5}, \textsc{gpt-4o}, and \textsc{llama-3.1}. Applying the Table A.\ref{table:criteria-generation-prompt} prompt with temperature=0 to facilitate deterministic output, each criteria writer generated one criteria set per fragment. We post-processed this output to identify each individual criterion according to its expected numerical label in the set. \S\ref{generated-criteria-descriptive-analysis} gives some descriptive analysis of the criteria sets. 

\subsection{Criteria-based Ratings}\label{section:rating_approach}

\begin{table*}[h!]
\small
\centering
\rowcolors{1}{gray!20}{white}
\begin{tabular}{ l | c c |c c |c c || c c }
\hiderowcolors
& \multicolumn{8}{c}{\textbf{VLM Rater}}\\
\cline{2-9}
\textbf{Criteria Writer} & \multicolumn{2}{c|}{\textbf{\textsc{claude-3.5}}} & \multicolumn{2}{c|}{\textbf{\textsc{gpt-4o}}} & \multicolumn{2}{c||}{\textbf{\textsc{pixtral}}} & \multicolumn{2}{c}{\textbf{$Average$}} \\
\hline
& Criterial & Base & Criterial & Base & Criterial & Base & Criterial & Base\\
\cline{2-9}
\showrowcolors
\textsc{claude-3.5} & 0.717 & 0.606 & 0.709 & 0.567 & 0.712 & 0.589 & 0.713 & 0.587 \\
\textsc{gpt-4o} & 0.701 & 0.602 & 0.687 & 0.583 & 0.699 & 0.589 & 0.695 & 0.592 \\
\textsc{llama-3.1} & 0.684 & 0.597 & 0.678 & 0.589 & 0.677 & 0.581 & 0.679 & 0.589 \\
\hline
\hline
\rowcolor{gray!40} $Average$ & 0.700 & 0.602 & 0.691 & 0.580 & 0.696 & 0.586 & 0.696 & 0.589\\
\hline
\end{tabular}
\caption{Accuracy of criterial and baseline (Base) raters grouped by criteria writer and VLM}
\label{table:criteria-scoring-accuracy}
\end{table*}

After obtaining the criteria sets, we then enlisted visually-enabled models to assess illustrations based on this criteria. In our scheme, when applying a criteria set to score a given illustration, each criterion receives a response indicating whether or not it is satisfied by the image. The overall illustration quality is quantified by the total number of satisfied criteria. Our scoring protocol is as follows: a response conveying that the criterion is satisfied is assigned 1.0 points; a response conveying ``maybe'' or partial satisfaction is assigned 0.5 points; and a response conveying the criterion is not satisfied is assigned 0.0 points. The total score for an illustration is the sum of these point values.

We implemented this by prompting a visually-enabled LLM (i.e. VLM) to assign responses to each criterion for a given illustration. We use the term \textit{criterial rater} to refer to a VLM's role when running this prompt, which appears in Table A.\ref{table:criteria-scoring-prompt}. As shown, the rater observes an illustration and the criteria set for the corresponding fragment. The rater is asked to respond to each criterion (where a response of `\cmark' means the criterion is satisfied, `\xmark' means not satisfied, and `?' means ``maybe''). As post-processing, we parsed these response tokens and mapped them to the point values defined above to obtain the illustration score. Table \ref{table:criteria-scoring-example} exemplifies this approach applied to both illustrations in a pair.

\paragraph{Rater Details} For raters, we utilized three VLMs that have obtained notable performance on visual understanding benchmarks: \textsc{claude-3.5}, \textsc{gpt-4o}, and \textsc{pixtral}\footnote{pixtral-large-2411, ran via the \href{https://console.mistral.ai/}{MistralAI API}} \cite{pixtral-large}. Each rater ran the prompt in Table A.\ref{table:criteria-scoring-prompt} with temperature=0. All images were resized to a height of 240 pixels with proportional width. We briefly assessed the correctness of raters' responses, which appears in \S\ref{rater-response-performance}.

\paragraph{Comparative Baseline} To determine the impact of criteria in assessing quality, we designed a comparable rating approach that scores illustrations on the same scale as the criterial rater but without observing the criteria itself. We use the term \textit{baseline rater} to refer to a VLM's application of the prompt for this approach, which is shown in Table A.\ref{table:baseline-scoring-prompt}. The prompt presents the fragment and illustration, and instructs the VLM to assign a rating in half-point increments between 0 and a maximum that is dynamically set to the length of the given criteria set. For each criteria writer, we compare the result obtained by a particular criterial rater to the analogous result obtained by the baseline rater.

\subsection{Selection Performance Results}

We applied all raters to score the illustrations in the \textit{Perfect-Agreement} subset of \textsc{SceneIllustrations}. For a given pair, a rater's selection was the image it assigned a higher score. We measured each rater's performance in terms of proportion of pairs where the rater's selection matched the human selection. We refer to this metric as \textit{accuracy}.

Table \ref{table:criteria-scoring-accuracy} shows the accuracy for all raters on these pairs, with the respective averages for each criteria writer and rater. For reference, always selecting the second illustration in each pair yields 49.4\% accuracy. We observe that the criterial raters all considerably outperform the baseline raters (an average accuracy of $\approx$70\% vs. 59\%). Criteria from different writers yields comparable results, with \textsc{claude-3.5} averaging the highest accuracy across raters ($\approx$71\%). The raters obtain similar accuracies when applied to the same criteria. Overall this outcome suggests that criteria are an effective strategy for modeling illustration quality, which in turn provides further evidence of LLMs' capacity to verbalize visual characteristics of narrative scenes. This leaves room for further accuracy gains, motivating future exploration of this dataset for understanding what makes a compelling scene illustration.





\section{Conclusion and Future Work}

This paper details a pipeline for generating illustrations of narrative scenes, which we apply to produce \textsc{SceneIllustrations}, a quality-annotated dataset of illustrations for a popular story corpus. We identify that LLMs can facilitate this illustration task by distilling scene descriptions from story text. We show that this capacity to verbalize implicit scene knowledge is also useful for modeling illustration quality. 

The scene illustration task isolates text-image alignment challenges in story visualization from issues of image-image alignment. In future work, we plan to consider recent approaches addressing the latter, such as ensuring visual consistency between story elements \cite[e.g.][]{liu2025onepromptonestory} and progressive story development across images \cite[e.g.][]{maharana2022storydalle}, in order to extend our illustration pipeline to generate multi-scene image sequences that depict complete stories.






\section*{Limitations}

We consider the following limitations:

\paragraph{Proprietary Models} Our scene illustration pipeline has a plug-and-play design, enabling any LLM to be used for fragmentation and scene captioning and any text-to-image model to be used for image generation. However, most of the models we assessed in this paper are proprietary (i.e. closed-weight), with exception to \textsc{llama-3.1} and \textsc{sd-3.5-large}. While the gap between closed and open-weight models is narrowing \cite{cottier2024how}, currently most models with capabilities relevant to the illustration task are closed-weight. This poses a general disadvantage in accessibility and reproducibility, which applies likewise to this work.

\paragraph{Prompt Design} Currently there is no tractable way to ensure that a particular prompt is optimal for the task it is intended to perform. Prompt optimization is fundamentally a process of iterative trial-and-error, even when automation is used to increase the number of trials. For our experiments, we primarily employed a principled approach to writing prompts, which involved adhering to general guidance on effective prompt design such as explaining instructions clearly and including representative exemplars \cite[e.g.][]{dairai2025general}. We iterated on this design according to qualitative subjective assessment of model outputs for inputs not included in our scene illustration dataset (i.e. ``vibe-based'' prompt engineering), rather than employing a quantitative optimization approach \cite[e.g.][]{khattab2024dspy} based on targets in a designated development set. There are tradeoffs to this technique: while it avoids overfitting to our presented dataset, it leaves open the possibility of further prompt optimization, which could yield a different view of model behavior compared with our observations.

\paragraph{Story Corpus} The story corpus we use, ROCStories, is popular in NLP research for some of the same reasons discussed in \S\ref{section:scene-illustration-dataset}: the constrained language and structure of the text makes the narrative elements more accessible to computational modeling techniques. The stories were authored specifically for the benefit of this research. However, this corpus is distinct from ``naturally'' authored stories whose complexity is what makes them compelling to readers. We have not yet fully assessed whether our scene illustration pipeline generalizes to more complex narratives.

\section*{Ethical Considerations}

Generative AI models, and in particular text-to-image models, pose various ethical risks \cite{bird2023typology}. A key risk is misinformation, which we avoid by utilizing stories that are strictly depictions of fictitious people and scenarios. We were primarily concerned with the risk of exposing Prolific annotators to harmful content. We attempted to mitigate this risk by manually reviewing stories sampled for inclusion in our dataset. We flagged stories that we anticipated could yield objectionable illustrations, and re-sampled a different story to replace each of these. Ultimately, this re-sampling was triggered for 10 stories. Of course, this procedure did not eliminate the risk, so we also utilized the content warning feature on the Prolific platform, which indicated to potential annotators that the task could expose them to offensive and/or biased content.

\bibliography{anthology,custom}

\clearpage
\appendix

\section{Appendix}

\subsection{Additional Statistics for \textsc{SceneIllustrations} Dataset}\label{additional-dataset-statistics}

\subsubsection{Analysis of Phase 1 Fragments}\label{phase-1-fragmentation-analysis}

As mentioned in \S\ref{section:phase-1-pipeline-details}, there were 206 total fragments derived from the 50 stories in Phase 1, based on applying \textsc{claude-3.5} to the prompt in Table \ref{table:fragmentation-prompt}. As shown in Table \ref{table:fragmentation-statistics}, the majority of fragments consist of a single sentence, with some consisting of 2 sentences and a few having 3 sentences. An internal annotator assessed each fragment to determine if it was the correctly-sized unit for a scene illustration. A fragment was considered incorrectly-sized if it either did not include all the text in the story relevant to a single scene (i.e. the fragment was too short) or if it included text pertaining to more than one scene (i.e. the fragment was too long). The annotator considered the vast majority of fragments to be correctly-sized ($\approx$96\%). 


\begin{table}[h!]
\small
\centering
\begin{tabular}{l l}
\hline
\# Total Fragments & 206 \\
\# 1-Sentence Fragments & 164 \\
\# 2-Sentence Fragments & 40 \\
\# 3-Sentence Fragments & 2 \\
\rowcolor{gray!20}Mean \# Sentences Per Fragment & 1.21\\
\rowcolor{gray!20}Mean \# Fragments Per Story & 4.12 \\
\rowcolor{gray!20}\% of Correctly-Sized Fragments & 96.1\% \\
\hline
\end{tabular}
\caption{Fragmentation statistics for stories in Phase 1}
\label{table:fragmentation-statistics}
\end{table}




\begin{table}[h!]
\small
\centering
\begin{tabular}{l c}
\textbf{Illustration Type} & \textbf{\# Illustrations}\\
\hline
\multicolumn{2}{c}{\underline{\textit{Phase 1}}}\\
All & 1576\\
\rowcolor{gray!20} \multicolumn{2}{l}{By Scene Description}\\
\quad\textsc{NC-Fragment} & 395\\
\quad\textsc{VC-Fragment} & 384\\
\quad\textsc{SC-Fragment} & 393\\
\quad\textsc{Caption} & 404\\
\rowcolor{gray!20} \multicolumn{2}{l}{By Image Generator} \\
\showrowcolors
\quad\textsc{flux-1-pro} & 791\\
\quad\textsc{mj-6.1} & 785\\
\hline
\multicolumn{2}{c}{\underline{\textit{Phase 2}}}\\
All & 1577\\
\rowcolor{gray!20} \multicolumn{2}{l}{By Scene Captioner}\\
\quad\textsc{claude-3.5} & 493\\
\quad\textsc{gpt-4o} & 531\\
\quad\textsc{llama-3.1} & 553\\
\rowcolor{gray!20} \multicolumn{2}{l}{By Image Generator} \\
\quad\textsc{flux-1.1-pro} & 307\\
\quad\textsc{ideogram-2.0} & 300\\
\quad\textsc{mj-6.1} & 318\\
\quad\textsc{recraft-v3} & 322\\
\quad\textsc{sd-3.5-large} & 330\\
\hline
\end{tabular}
\caption{Number of unique illustrations associated with each scene description type and image generator in Phase 1 and Phase 2}
\label{table:unique-illustration-statistics}
\end{table}

\subsubsection{Win Rates for Image Generators}\label{win-rates-image-generators}

To determine whether the choice of image generator influenced illustration quality in both Phase 1 and Phase 2, we computed the win rates for each image generator against each other among the pairs that used different image generators. 

For Phase 1, there were only two image generators used to produce illustrations, \textsc{flux-1-pro} vs. \textsc{mj-6.1}. We did not find any significant difference in their win rates. Table \ref{table:annotation-task-1-image-generator-win-stats} shows these results.

\begin{table}[h!]
\small
\centering
\begin{tabular}{l l}
\textbf{\textsc{flux-1-pro}} & \textbf{\textsc{mj-6.1}}\\
\hline
42.6\% & 41.0\%\\
\end{tabular}
\caption{Win rates (percentages) of \textsc{flux-1-pro} vs \textsc{mj-6.1} for Phase 1 pairs}.
\label{table:annotation-task-1-image-generator-win-stats}
\end{table}

The Phase 2 data utilized a larger set of image generators. Table \ref{table:annotation-task-2-image-generator-win-stats} shows their win rates, presented comparably to Table \ref{table:annotation-task-2-scene-captioner-win-stats} where each value is the percentage of selections for the image generator in the row against the image generator in the column. According to these results, \textsc{ideogram-2.0} obtains the highest win rates against the other image generators, with significant success against \textsc{flux-1.1-pro}, \textsc{mj-6.1}, and \textsc{sd-3.5-large}. Additionally, \textsc{recraft-v3} is significantly favored over \textsc{mj-6.1}. Analysis of these model differences for this task is an opportunity for future work.

\begin{table*}[h!]
\small
\centering
\rowcolors{1}{white}{gray!20}
\begin{tabular}{l|c|c|c|c|c}
& \textsc{flux-1.1-pro} & \textsc{ideogram-2.0} & \textsc{mj-6.1} & \textsc{recraft-v3} & \textsc{sd-3.5-large} \\
\hline
\textsc{flux-1.1-pro} & - & 35.4 & 43.2\phantom{*} & 45.3 & 48.6\phantom{*} \\
\textsc{ideogram-2.0} & 53.5* & - & 61.7* & 46.9 & 58.5* \\
\textsc{mj-6.1} & 39.9\phantom{*} & 32.1 & - & 29.2 & 45.4\phantom{*} \\
\textsc{recraft-v3} & 44.0\phantom{*} & 40.1 & 61.8* & - & 50.6\phantom{*} \\
\textsc{sd-3.5-large} & 43.7\phantom{*} & 30.3 & 43.7\phantom{*} & 37.8 & - \\
\hline
\end{tabular}
\caption{Win rates (percentages) by image generator for Phase 2. Statistically significant win rates are denoted with an asterisk.}
\label{table:annotation-task-2-image-generator-win-stats}
\end{table*}

\subsection{Criteria-based Evaluation Details}\label{criteria-generation}



\subsubsection{Descriptive Analysis of Criteria Sets}\label{generated-criteria-descriptive-analysis}

Regarding the generated criteria sets (\S\ref{section:criteria-generation-approach}), Table \ref{table:mean-criteria-set-length} compares the average number of criteria in the sets generated by each criteria writer, revealing that \textsc{claude-3.5} generated the longest criteria sets, followed by \textsc{gpt-4o}, and \textsc{llama-3.1}. 

\begin{table}[H]
\small
\centering
\begin{tabular}{l|l|l}
\textbf{\textsc{claude-3.5}} & \textbf{\textsc{gpt-4o}} & \textbf{\textsc{llama-3.1}} \\
\hline
19.3 & 17.3 & 15.8\\
\end{tabular}
\caption{Mean number of criteria per set for each writer}
\label{table:mean-criteria-set-length}
\end{table}

\begin{figure}[h!]
    \centering
    \includegraphics[width=\linewidth]{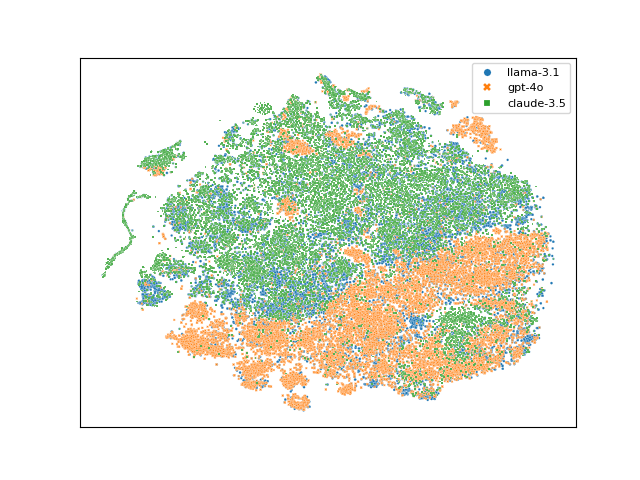}
    \caption{Visualization of criteria generated by each criteria writer. Each point is a single criterion represented by its ModernBERT embedding. We applied PCA followed by t-SNE to plot the embeddings in 2D space.}
    \label{figure:embedded-criteria-plot}
\end{figure}

Additionally, Figure \ref{figure:embedded-criteria-plot} visualizes all criteria, based on encoding each criterion with the ModernBERT embedding model \cite{warner2024smarterbetterfasterlonger}, then running PCA + t-SNE to yield a 2D embedding. While there are no distinct clusters associated with each criteria writer, some separation can be observed between the criteria generated by \textsc{claude-3.5} and \textsc{gpt-4o}, while those generated by \textsc{llama-3.1} are more distributed alongside both other criteria writers.

\subsubsection{Criterial Rater Assessment}\label{rater-response-performance}

\begin{table}[h!]
\small
\centering
\rowcolors{1}{white}{gray!20}
\begin{tabular}{l|l}
\textbf{Rater} & \textbf{$\kappa$}\\
\hline
\textsc{claude-3.5} & 0.676\\
\textsc{gpt-4o} & 0.710\\
\textsc{pixtral} & 0.622\\
\hline
\end{tabular}
\caption{Correctness of criterial rater responses ($\kappa$)}
\label{table:criteria-scoring-gold-kappa}
\end{table}

As referenced in \S \ref{section:rating_approach}, we conducted a small assessment of the correctness of the VLM raters' responses to criteria. To do this, we randomly sampled 100 items, each with a unique image and criteria set. We then enlisted an expert human annotator to assign a response to each criterion, which we treated as the gold standard criterion response for the sampled image. We measured rater correctness in terms of linear-weighted $\kappa$ agreement with the gold standard, where responses of `\xmark' were mapped to -1, `?' to 0, and `\cmark' to 1; this results in less weight assigned to disagreements involving `?' (``maybe'') responses. Table \ref{table:criteria-scoring-gold-kappa} shows the $\kappa$ on these 1699 criterion responses. It indicates that raters are all substantially aligned with the human annotator, though \textsc{gpt-4o} appears to have the highest human agreement, followed by \textsc{claude-3.5}, and then \textsc{pixtral}.

\begin{table*}[h]
\small
\begin{tabularx}{\textwidth}{X}
\hline
\cellcolor{instructioncolor}You are performing the task of story fragmentation. The task is to split a story into fragments where each fragment consists of a distinct part of the story. A fragment contains enough information to yield a visualization that is unique to that part of the story. In this version of the task, you will insert brackets (i.e. [ and ]) into the given story text to annotate the beginning and end of each fragment. Write the fragments without preamble. Here are some examples:\\
\vspace{0.1cm}\\
\cellcolor{exemplarcolor}Story: Mia sat at home in her living room watching sports. Her favorite soccer team was playing their rival. To encourage her team, she began chanting positive phrases. During her chant, her favorite team scored a goal. Mia cheered loudly and thought that she helped score that goal.\newline
Fragmented Story: [Mia sat at home in her living room watching sports. Her favorite soccer team was playing their rival.] [To encourage her team, she began chanting positive phrases.] [During her chant, her favorite team scored a goal.] [Mia cheered loudly and thought that she helped score that goal.]\newline\newline
\textit{[...2 more exemplars...]}\\
\vspace{0.1cm}\\
Story: \texttt{\{\{story\}\}}\newline
Fragmented Story:\\
\hline
\end{tabularx}
\caption{\textbf{Fragmentation prompt}. LLM prompt for annotating fragment boundaries in a story, which consists of a \colorbox{instructioncolor}{task instruction} and \colorbox{exemplarcolor}{exemplars} demonstrating the task. The stories in the exemplars are from various corpora (\href{https://cs.rochester.edu/nlp/rocstories/}{ROCStories}, \href{https://huggingface.co/datasets/roneneldan/TinyStories}{TinyStories}, and the \href{https://github.com/USArmyResearchLab/ARL-Creative-Visual-Storytelling}{ARL Creative Visual Storytelling Anthology}).}
\label{table:fragmentation-prompt}
\end{table*}
\label{sec:appendix}

\begin{table*}[h]
\small
\begin{tabularx}{\textwidth}{X}
\hline
\cellcolor{instructioncolor}You will be shown a story fragment along with its story context. Your task is to rewrite the fragment so that its meaning can be fully understood if read independently of the story context. For instance, you should replace names of characters with generic nouns. You should replace pronouns with the nouns they refer to (if the reference is a character, replace it with the appropriate generic noun). For first-person pronouns, replace the pronoun with a generic identifier (e.g. "I" -> "A person", "my" -> "the person's"). If the fragment implicitly refers to any other information in the story context, this information should be made explicit in the revised fragment. Write the revised fragment without preamble. Here are some examples:\\
\vspace{0.1cm}\\

\cellcolor{exemplarcolor}Story Context: Anna was filling her bird feeders. But a chunk of suet fell onto the ground. Her dog rushed over and lapped it up! Anna was astonished. She had no idea dogs loved bird food!\newline
Story Fragment: Her dog rushed over and lapped it up!\newline
Revised Story Fragment: The woman's dog rushed over and lapped up the chunk of suet that had fallen onto the ground.\newline\newline
\textit{[...2 more exemplars...]}\\
\vspace{0.1cm}\\
Story Context: \texttt{\{\{story\}\}}\newline
Story Fragment: \texttt{\{\{fragment\}\}}\newline
Revised Story Fragment:\\
\hline
\end{tabularx}
\caption{\textbf{Fragment rewriting prompt}. LLM prompt for generating \textsc{SC-Fragment} scene descriptions. The prompt consists of a \colorbox{instructioncolor}{task instruction} and \colorbox{exemplarcolor}{exemplars} demonstrating the task. The stories in the exemplars are from the \href{https://cs.rochester.edu/nlp/rocstories/}{ROCStories} corpus.}
\label{table:fragment-rewriting-prompt}
\end{table*}

\begin{table*}[h]
\small
\begin{tabularx}{\textwidth}{X}
\hline
\cellcolor{instructioncolor}Imagine an AI system will be used to generate illustrations for story fragments. This AI illustrator generates a single image given a caption describing what is contained in the image. Your task is to read a story fragment along with its story context and write a caption that describes how to illustrate the fragment. The caption should elaborately describe the most salient way to visualize the fragment. It should completely specify all the information the illustrator needs to generate the image. Write the caption without preamble. Here are some examples: \\
\vspace{0.1cm}\\

\cellcolor{exemplarcolor}Story Context: Carrie had just learned how to ride a bike. She didn't have a bike of her own. Carrie would sneak rides on her sister's bike. She got nervous on a hill and crashed into a wall. The bike frame bent and Carrie got a deep gash on her leg.\newline
Story Fragment: Carrie would sneak rides on her sister's bike.\newline
Caption for Story Fragment: A young girl with a mischievous expression carefully wheels a bicycle that's slightly too big for her out of a garage, glancing over her shoulder as if making sure no one sees her.\newline\newline
\textit{[...2 more exemplars...]}\\
\vspace{0.1cm}\\
Story Context: \texttt{\{\{story\}\}}\newline
Story Fragment: \texttt{\{\{fragment\}\}}\newline
Caption for Story Fragment:\\
\hline
\end{tabularx}
\caption{\textbf{Scene captioning prompt}. LLM prompt for generating \textsc{Caption} scene descriptions. The prompt consists of a \colorbox{instructioncolor}{task instruction} and \colorbox{exemplarcolor}{exemplars} demonstrating the task. The stories in the exemplars are from various corpora (\href{https://cs.rochester.edu/nlp/rocstories/}{ROCStories}, \href{https://huggingface.co/datasets/roneneldan/TinyStories}{TinyStories}, and the \href{https://github.com/USArmyResearchLab/ARL-Creative-Visual-Storytelling}{ARL Creative Visual Storytelling Anthology}).}
\label{table:scene-captioning-prompt}
\end{table*}

\begin{table*}
\small
\rowcolors{1}{white}{gray!20}
\begin{tabular}{ m{0.2\textwidth} m{0.3\textwidth} m{0.2\textwidth} m{0.2\textwidth}}
\textbf{Fragment} (within story) & \textbf{\textsc{Caption}} &  \textbf{ \textsc{Caption} Illustration} & \textbf{\textsc{VC-Fragment}\newline Illustration}\\
\hline
\textbf{Joe went to the junkyard to find treasures.} He dug through a stack of old clothes. At the bottom was a vintage shirt with a rock band on it. Joe put the shirt on to try it out. Joe was happy the shirt fit. & A young man with an excited expression stands at the entrance of a sprawling junkyard. He's wearing casual clothes and holding a small backpack. The junkyard is filled with piles of discarded items, including old appliances, furniture, and scrap metal. The sky is bright and sunny, creating an adventurous atmosphere. In the foreground, a rusty sign reads "Junkyard" with operating hours listed below. & \includegraphics[width=0.2\textwidth]{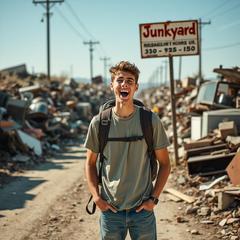} & \includegraphics[width=0.2\textwidth]{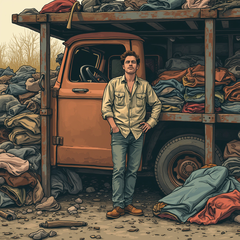}\\
Trish hated the outdoors. Her friends convinced her to go camping. \textbf{She wasn't having a good time.} They showed her how to fish and showed her the stars. Eventually Trish grudgingly came to accept nature. & A young woman with a scowling expression sits on a log by a campfire, arms crossed and looking miserable. She's surrounded by cheerful friends setting up tents and unpacking camping gear in a forest clearing. Her clean, urban clothing contrasts with the rugged outdoor setting, emphasizing her discomfort with nature. & \includegraphics[width=0.2\textwidth]{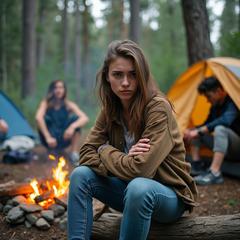} & \includegraphics[width=0.2\textwidth]{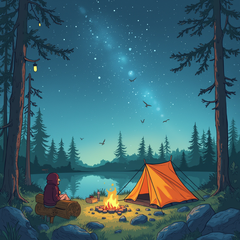}\\
Sammy's coffee grinder was broken. He needed something to crush up his coffee beans. He put his coffee beans in a plastic bag. \textbf{He tried crushing them with a hammer.} It worked for Sammy. & A man in casual clothing stands at a kitchen counter, holding a hammer above a clear plastic bag filled with whole coffee beans. The hammer is poised mid-swing, about to strike the bag. The man's face shows a mix of determination and uncertainty. Scattered around the counter are a few stray coffee beans and an unplugged, visibly broken coffee grinder. & \includegraphics[width=0.2\textwidth]{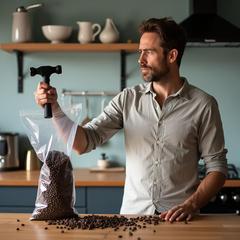} & \includegraphics[width=0.2\textwidth]{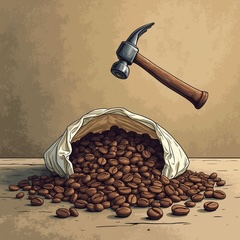}\\
I decided to go on a bike ride with my brother. We both headed out in the morning. We were having a lot of fun. Suddenly, he hit a rock and broke his wheel! \textbf{I felt very badly for my brother}. & A concerned young person stands next to their brother, who sits dejectedly on the ground next to a fallen bicycle with a visibly bent front wheel. The scene takes place on a sunny morning on a bike path, with trees and nature in the background. The standing sibling has a sympathetic expression, reaching out to comfort their brother, who looks disappointed and upset about the broken bike. & \includegraphics[width=0.2\textwidth]{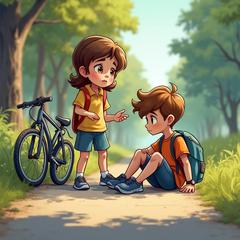} & \includegraphics[width=0.2\textwidth]{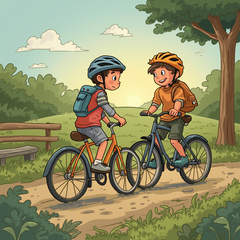}\\
\hline
\end{tabular}
\caption{Examples of scene illustrations in Phase 1. For each story fragment, we show an illustration resulting from the LLM-generated \textsc{Caption} scene description and one resulting from the baseline \textsc{VC-Fragment} scene description. The image generator for all illustrations is \textsc{flux-1-pro}.}
\label{table:illustration-examples-phase-1}
\end{table*}

\begin{figure*}[h]
    \centering
    \includegraphics[width=\textwidth]{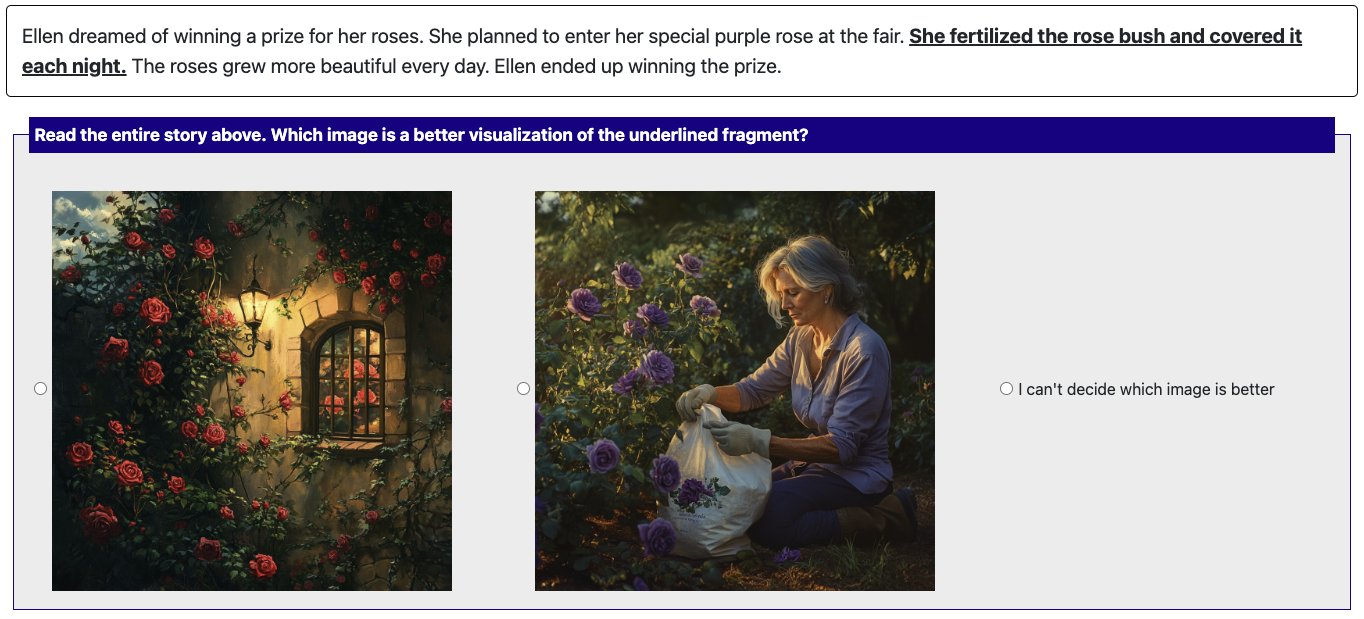}
    \caption{Example of a item shown to participants in the annotation task described in \S \ref{section:phase-1-annotation-task}}
    \label{figure:annotation-interface}
\end{figure*}

\begin{table*}[h!]
\small
\centering
\begin{tabularx}{0.925\textwidth}{X | p{0.185\textwidth} | p{0.12\textwidth} | p{0.15\textwidth}}
& \multicolumn{3}{c}{\textbf{\textsc{Caption} Win \%}}\\
\cline{2-4}
\textbf{Scene Description Pair} & \textsc{mj-6.1} \& \textsc{flux-1-pro} & \textsc{mj-6.1} Only & \textsc{flux-1-pro} Only \\
\hline
\rowcolor{gray!20}\textsc{Caption} vs. \textsc{NC-Fragment}  & 78.1  & 79.2 & 77.7 \\
\textsc{Caption} vs. \textsc{VC-Fragment} & 74.7 & 74.5 & 75.0\\
\rowcolor{gray!20}\textsc{Caption} vs. \textsc{SC-Fragment} & 72.5 & 76.1 & 68.9\\
\hline
\end{tabularx}
\caption{Extended view of Table \ref{table:annotation-task-1-scene-description-win-rates}. Here, the win rates (percentages) for \textsc{Caption} vs. baseline scene descriptions in Phase 1 are split out by pairs where both illustrations used the same image generator (the \textsc{mj-6.1} Only and \textsc{flux-1-pro} Only columns). This shows that the \textsc{Caption} win rate is similar regardless of which image generator is used.}
\label{table:annotation-task-1-scene-description-win-rates-extended}
\end{table*}

\begin{table*}[h]
\small
\begin{tabularx}{\textwidth}{X}
\hline
\cellcolor{instructioncolor}Imagine an AI system will be used to judge the quality of images intended to illustrate story fragments. This AI judge scores the images given some criteria about what should be depicted in the images. Your task involves writing the criteria for this AI judge. In particular, you will read a story and focus on a fragment at a specific position in the story. You will write the criteria defining the characteristics the image for that fragment should satisfy in order to be considered a good illustration of the fragment. There are a few things to consider when writing the criteria. First, while the criteria should define the fundamental characteristics depicted in the image, the visual details of these characteristics may vary across images, and alternative details may be similarly effective in illustrating the fragment. Each criterion should be written in a way that accommodates these potential variations in detail, without assuming specific information that is not defined explicitly in the story. Additionally, each criterion should refer to only a single atomic characteristic of the image. If a criterion references multiple characteristics such that an image might satisfy some but not others, it should be further split into multiple separate criteria. For example, instead of writing "the image shows a sapphire ring on the bathroom floor" as one criterion, you should write "the image shows a ring", "the ring contains a sapphire", and "the ring is on the bathroom floor" as separate criteria. The criteria should not only consider the presence of certain elements in the image, but also the visual quality of their depiction. Write the criteria without preamble, with a number header (e.g. '1.') for each criterion. Try to write as many criteria as possible, but avoid specifying extraneous or redundant criteria. Here is an example:\\
\vspace{0.1cm}\\
\cellcolor{exemplarcolor}Story Context: Lisa has a beautiful sapphire ring. She always takes it off to wash her hands. One afternoon, she noticed it was missing from her finger! Lisa searched everywhere she had been that day. She was elated when she found it on the bathroom floor!\newline
Story Fragment: She was elated when she found it on the bathroom floor!\newline
Image Criteria for Story Fragment:\newline
1. The image shows a clearly visible ring \newline
2. The image portrays a bathroom setting recognizable through typical bathroom elements (tiles, fixtures, etc.)\newline
3. The ring contains a blue gemstone recognizable as a sapphire\newline
4. The ring is on the bathroom floor\newline
5. The ring appears to be positioned naturally as if it had fallen or been dropped\newline
6. A female figure (Lisa) is present in the image\newline
7. The woman's facial expression clearly conveys joy or elation\newline
8. The woman's body language demonstrates excitement or relief\newline
9. The woman's positioning suggests she has just discovered or is reaching for the ring\newline
10. The lighting adequately illuminates the ring to make it visible as the focal point\newline
11. The perspective of the image allows viewers to see both the ring and the woman's emotional reaction\newline
12. The composition draws attention to the moment of discovery\newline
13. The spatial relationship between the woman and ring suggests imminent retrieval\newline
14. The overall scene composition captures the spontaneous nature of the discovery\newline
15. The woman's appearance suggests this is taking place during daytime/afternoon\newline
16. The ring appears intact and undamaged, justifying the woman's relief\newline
17. The bathroom setting appears residential rather than public\\
\vspace{0.1cm}\\

Story Context: \texttt{\{\{story\}\}}\newline
Story Fragment: \texttt{\{\{fragment\}\}}\newline
Image Criteria for Story Fragment:\\
\hline
\end{tabularx}
\caption{\textbf{Criteria generation prompt}. LLM prompt used to generate evaluation criteria for assessing the quality of scene illustrations.}
\label{table:criteria-generation-prompt}
\end{table*}

\begin{table*}[h]
\small
\begin{tabularx}{\textwidth}{X}
\hline
\cellcolor{instructioncolor}You will observe an image along with a list of criteria, where each criterion describes a characteristic or quality that may or may not be depicted in the image. Your task is to determine whether or not each criterion is satisfied by the image. For each criterion, if the image fully satisfies that criterion, write a checkmark ('\cmark') after it. If the image only partially satisfies the criterion but not completely, write a question mark ('?') after it. Otherwise, if the image does not satisfy that criterion, write an X mark ('\xmark') after it.  Reiterate each criterion before giving your assessment for it, but do not provide additional preamble in your response. Here is an example:\\
\vspace{01.cm}\\
\cellcolor{exemplarcolor}Criteria:\newline
1. The image shows a young woman (Laura) in an apartment setting\newline
2. The woman's facial expression conveys happiness or contentment\newline
3. The apartment appears to be newly moved into, with some visible unpacked items\newline
4. There are visible windows in the apartment\newline
5. The view through the windows shows recognizable California scenery (palm trees, ocean, mountains, or urban landscape)\newline
6. The lighting suggests natural daylight entering the apartment\newline
7. The apartment appears residential and suitable for a recent college graduate\newline
Image: <IMAGE WILL APPEAR HERE>\newline
Criteria Responses:\newline
1. The image shows a young woman (Laura) in an apartment setting \cmark\newline
2. The woman's facial expression conveys happiness or contentment \xmark\newline
3. The apartment appears to be newly moved into, with some visible unpacked items ?\newline
4. There are visible windows in the apartment \cmark\newline
5. The view through the windows shows recognizable California scenery (palm trees, ocean, mountains, or urban landscape) \xmark\newline
6. The lighting suggests natural daylight entering the apartment \cmark\newline
7. The apartment appears residential and suitable for a recent college graduate \cmark\\
\vspace{0.1cm}\\
Criteria:\newline
\texttt{\{\{criteria\}\}}\newline
Image: \texttt{\{\{image\}\}}\newline
Criteria Responses:\\
\hline
\end{tabularx}
\caption{\textbf{Criteria-based rating prompt}. VLM prompt used to score the quality of a scene illustration by assigning responses to each criterion in a provided criteria set}
\label{table:criteria-scoring-prompt}
\end{table*}

\begin{table*}[h]
\small
\begin{tabularx}{\textwidth}{X}
\hline
\cellcolor{instructioncolor}Your task is to rate how well a particular image illustrates a fragment of a story. You will observe the fragment with its story context, alongside the image depicting the fragment. Provide a rating on a scale ranging from 0.0 to \texttt{\{\{len(criteria)\}\}} in half-point increments, where 0.0 indicates the image is unrelated to the fragment and \texttt{\{\{len(criteria)\}\}} indicates the image is a perfect illustration of the fragment. Do not provide additional preamble before the rating. Here is an example:\\
\vspace{01.cm}\\
\cellcolor{exemplarcolor}Story: Laura had just graduated college. She was planning on moving on California. She packed all her belongings in her car and drove 18 hours. When she arrived at her new apartment she unpacked all her things. Laura loved the new change of scenery at her new place.\newline
Fragment: Laura loved the new change of scenery at her new place.\newline
Image: <IMAGE WILL APPEAR HERE>\newline
Rating: 4.5\\
\vspace{0.1cm}\\
Story: \texttt{\{\{story\}\}}\newline
Fragment: \texttt{\{\{fragment\}\}}\newline
Image: \texttt{\{\{image\}\}}\newline
Rating:\\
\hline
\end{tabularx}
\caption{\textbf{Baseline rating prompt}. VLM prompt used to score the quality of a scene illustration by directly assigning a rating between 0 and a maximum. This maximum is dynamically set to the total number of criteria in a provided criteria set (\texttt{\{\{len(criteria)\}\}}), even though the criteria themselves are not referenced in the prompt.}
\label{table:baseline-scoring-prompt}
\end{table*}

\end{document}